\documentclass[letterpaper]{article} 
\usepackage{aaai23}  
\usepackage{times}  
\usepackage{helvet}  
\usepackage{courier}  
\usepackage[hyphens]{url}  
\usepackage{graphicx} 
\usepackage{natbib}  
\usepackage{caption} 
\usepackage{algorithm}
\usepackage{algorithmic}

\usepackage{newfloat}
\usepackage{listings}
\DeclareCaptionStyle{ruled}{labelfont=normalfont,labelsep=colon,strut=off} 
\floatstyle{ruled}
\newfloat{listing}{tb}{lst}{}
\floatname{listing}{Listing}
\usepackage{multirow}
\usepackage{enumitem}
\usepackage{graphics}
\usepackage{subcaption}
\usepackage{amsmath}
\usepackage{amsfonts}
\usepackage{diagbox}

\usepackage{xcolor}

\usepackage{pgfplots} 

\title{Crowd-level Abnormal Behavior Detection via  Multi-scale \\ Motion Consistency Learning}
\author{
    Linbo Luo\textsuperscript{\rm 1}\thanks{Corresponding author: lbluo@xidian.edu.cn},
    Yuanjing Li\textsuperscript{\rm 1},
    Haiyan Yin\textsuperscript{\rm 2},
    Shangwei Xie\textsuperscript{\rm 1},
    Ruimin Hu\textsuperscript{\rm 1},
    Wentong Cai\textsuperscript{\rm 3}
}
\affiliations{
    \textsuperscript{\rm 1} School of Cyber Engineering, Xidian University,
    \textsuperscript{\rm 2} Sea AI Lab\\
    \textsuperscript{\rm 3} School of Computer Science and Engineering, Nanyang Technological University\\

}

\begin{document}

\maketitle

\begin{abstract}
Detecting abnormal crowd motion emerging from complex interactions of individuals is paramount to ensure the safety of crowds. Crowd-level abnormal behaviors (CABs), e.g., counter flow and crowd turbulence, are proven to be the crucial causes of many crowd disasters. In the recent decade, video anomaly detection (VAD) techniques have achieved remarkable success in detecting individual-level abnormal behaviors (e.g., sudden running, fighting and stealing), but research on VAD for CABs is rather limited. Unlike individual-level anomaly, CABs usually do not exhibit salient difference from the normal behaviors when observed locally, and the scale of CABs could vary from one scenario to another. In this paper, we present a systematic study to tackle the important problem of VAD for CABs with a novel crowd motion learning framework, \emph{multi-scale motion consistency network} (MSMC-Net). MSMC-Net first captures the spatial and temporal crowd motion consistency information in a graph representation. Then, it simultaneously trains multiple feature graphs constructed at different scales to capture rich crowd patterns. An attention network is used to adaptively fuse the multi-scale features for better CAB detection. For the empirical study, we consider three large-scale crowd event datasets, UMN, Hajj and Love Parade. Experimental results show that MSMC-Net could substantially improve the state-of-the-art performance on all the datasets.
\end{abstract}

\section{Introduction}

In real-world crowd events, many self-organizing crowd behaviors could emerge from the complex interactions of individuals, where some of those behaviors, such as counter flow, stop-and-go waves and crowd turbulence, can bring excessive contact forces, making people lose balance, get crushed, or even suffocated~\cite{helbing2012crowd,li2020experimental}. Such hazardous self-organizing behaviors are often referred to as crowd-level abnormal behaviors (CABs). Many post-disaster analyses have revealed that CABs are the causes of fatalities in many crowd disasters~\cite{helbing2007dynamics,helbing2012crowd,helbing2013globally,ma2013new,zhao2020assessing}. In the past decade, more than ten thousand people have lost their lives or got injured in the  crowd disasters recorded  worldwide~\cite{wikilist}. Thus, it is paramount to effectively detect these CABs during large-scale crowd events to reduce crowd risks.

\begin{figure}[!t]
\setlength{\abovecaptionskip}{+0.9cm}
\setlength{\belowcaptionskip}{-0.5cm}
\centering
    \begin{subfigure}[t]{0.22\textwidth}
           \centering
           \includegraphics[width=\textwidth]{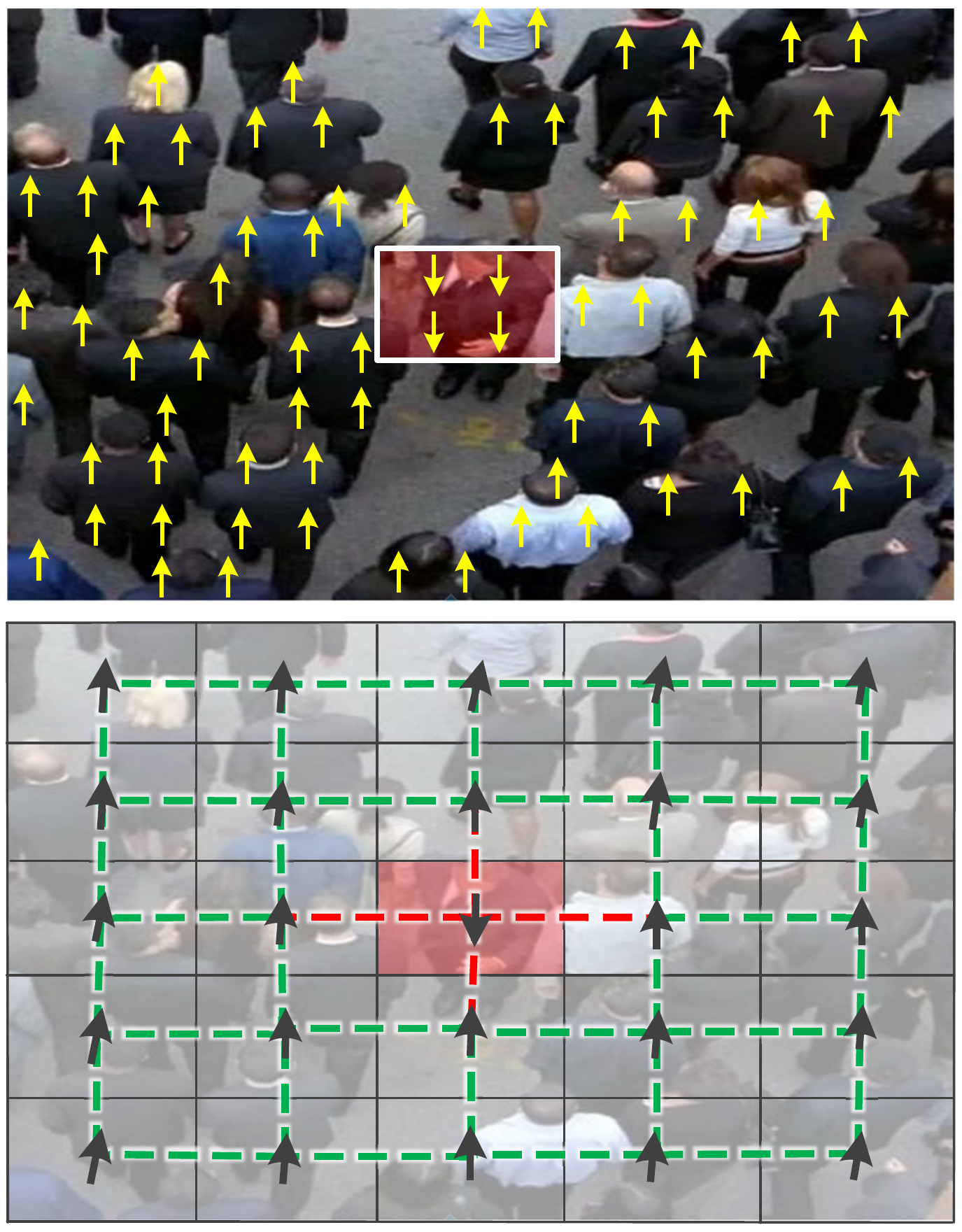}
            \caption{Individual-level anomaly}
            \label{fig:1a}
    \end{subfigure}\hskip 0.13in
    \begin{subfigure}[t]{0.22\textwidth}
            \centering
            \includegraphics[width=\textwidth]{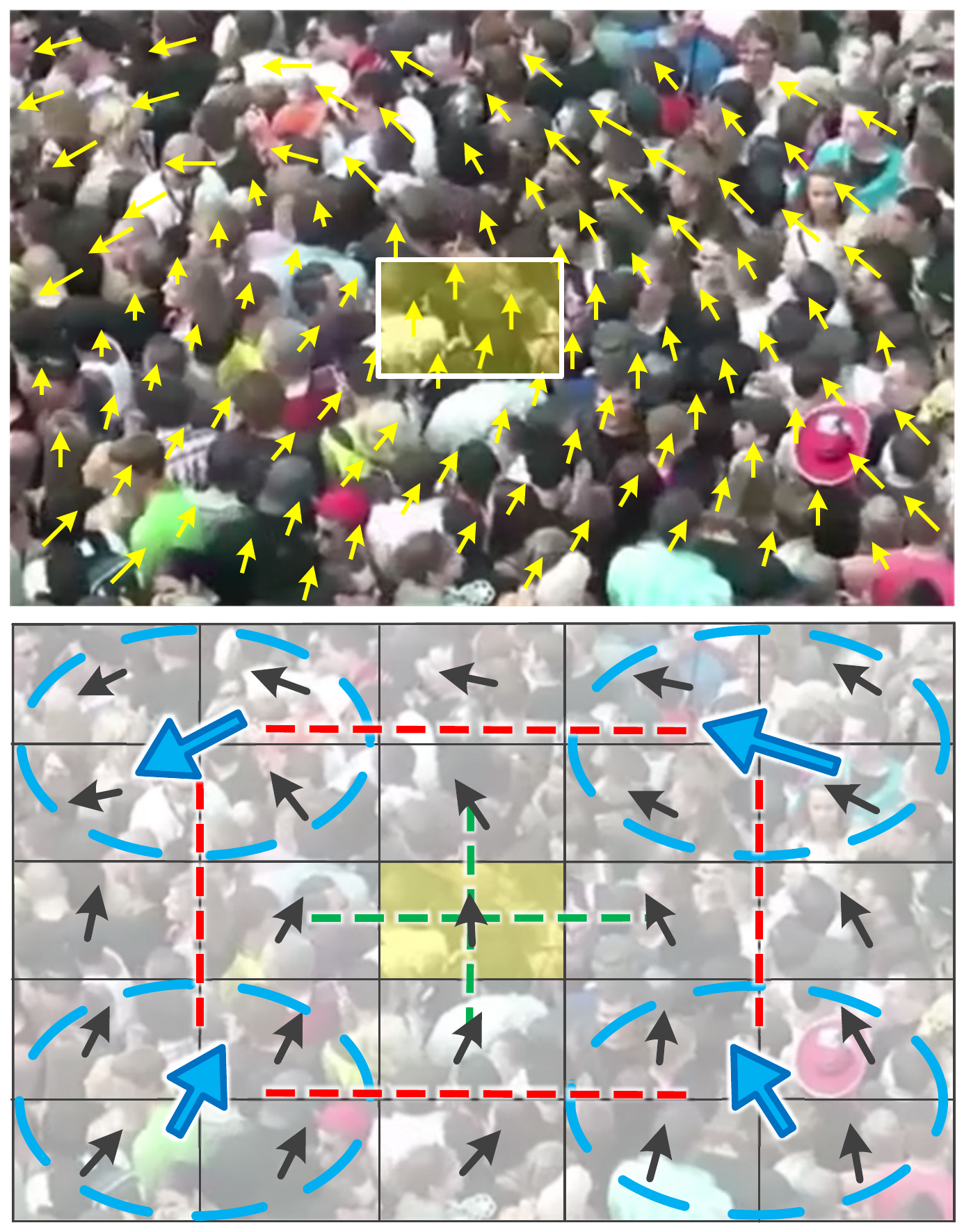}
            \caption{Crowd-level anomaly}
            \label{fig:1b}
    \end{subfigure}\vskip -0.1in
    \caption{An illustrative example of individual-level vs. crowd-level abnormal behaviors: (a) \textbf{individual-level} counter-direction pedestrian, whose motion in the local region (red rectangle) exhibits salient difference from neighbouring regions; (b) \textbf{crowd-level} turbulence, in which individuals' motions in the local region (yellow rectangle) do not differ much from neighbouring regions. (\textbf{Legend}: yellow arrows represent optical flow fields; black/blue arrows show the average velocity in each unit-region/circled-region; red/green dashed lines imply low/high motion consistency.)
}
\label{fig1}
\end{figure}

In recent decades, video anomaly detection (VAD) research has gained tremendous momentum. The task of VAD is to detect behaviors or events that are rare and/or have significantly different characteristics from the normal ones in videos. Despite the remarkable success achieved by the existing VAD methods, most of them are designed and testified for detecting individual-level abnormal behaviors, such as sudden running, fighting and stealing, thanks to the full range of publicly available datasets of this kind (e.g., UCSD~\cite{mahadevan2010anomaly}, ShanghaiTech~\cite{liu2018future}, UCF Crime~\cite{sultani2018real}, etc.). However, VAD for CABs is still very limited. We argue that it is challenging to directly apply the existing VAD methods to detect CABs due to the intrinsic difference between individual-level and crowd-level behaviors. First, crowd behavior patterns emerge in crowd motion at macroscopic level~\cite{helbing2013globally}. Unlike individual-level behaviors whose anomaly could be distinguished by the appearance or motion of abnormal individual(s) in a local region of a crowd, crowd-level anomaly does not always exhibit salient differences from normal ones when observed locally (see an example in Figure~\ref{fig1}). Thus, the existing VAD models, which are mostly designed to distinguish anomaly patterns by learning features related to local appearance (e.g., raising arm) or local motion (e.g., an unusual acceleration), are insufficient for detecting CABs. Second, the scale of crowd-level behaviors can vary more considerably under different environmental conditions and crowd densities~\cite{solmaz2012identifying} as compared to individual-level behaviors, which often have a relatively unified scale (e.g., in the range of one or a small group of pedestrians). Therefore, developing the VAD model against the varying scale of CABs is crucial to ensure robust performance of the CAB detection.

To the best of our knowledge, this paper is the first work that formally considers bridging deep learning-based VAD techniques and the CAB detection tasks. To tackle the aforementioned challenges of CAB detection, our work has the following two distinguishing properties. First, instead of detecting anomalies locally like conventional VAD methods, we advocate the analysis of global patterns of collective crowd motion to distinguish CABs from normal behaviors. To this end, we consider the modeling of \textit{crowd motion consistency}, an informative feature to quantify the collectiveness of crowd~\cite{zhou2013measuring,li2020quantifying}, for CAB detection tasks. Specifically, we introduce a graph-based crowd motion consistency representation, which aims to capture both spatial and time-varying characteristics of crowd motion based on the optical flows extracted from videos. Second, to make the detection robust to the varying scales of the CABs, we design a novel multi-scale motion learning framework, where the model could receive rich crowd behavior information from various feature graphs extracted under different scales for pattern recognition. We also introduce an attentional decoding module to effectively synthesize the multi-scale feature graphs for learning high-quality features for CAB~detection.

In a nutshell, our paper has the following contributions:
\begin{enumerate}[label=(\roman*)]
\item We motivate and introduce the first work that aims to develop a VAD method specifically for tackling the detection of large-scale crowd-level abnormal behaviors.
\item We propose a novel graph-based method, \emph{multi-scale motion consistency network} (MSMC-Net), which comes with a \emph{crowd motion consistency} representation learning module to capture both spatial and temporal motion consistency, as well as a multi-scale decoding module that leverages multiple feature graphs at different scales to capture the crowd behaviors with varying scales.
\item We present an extensive empirical evaluation study, where we implement five related baselines, adopt three datasets and demonstrate that our method leads to superior performance consistently across all the datasets.
\end{enumerate}

\section{Related Work}


\noindent\textbf{Video Anomaly Detection.} Given the ambiguity and diversity of abnormal behaviors in crowd videos, the mainstream of the current VAD research is the data-driven approach, which learns the normal behavior patterns in the training phase using only normal data and detecting the abnormal behavior in the testing phase by evaluating the deviation from the normal patterns. Most existing deep learning methods fall into either reconstruction-based~\cite{nguyen2019anomaly,zhou2019anomalynet,park2020learning} or prediction-based~\cite{liu2018future,ye2019anopcn,cai2021appearance,chen2022comprehensive} strands, in which often testing samples with high reconstruction or prediction errors are regarded as anomalies. More recently, the weakly-supervised approach of adding the video-level anomaly label data for training has also gained popularity~\cite{zhong2019graph,feng2021mist,li2022self}. In this work, we adopt the unsupervised learning approach for detecting CABs. The weakly-supervised learning approach is not considered because crowd-level abnormal behaviors occur less frequently (usually during crowd disasters) than individual-level anomalies, and the crowd anomaly data is thus too scarce to be sufficiently utilized for training. Among recent unsupervised VAD methods~\cite{cai2021appearance,chen2022comprehensive}, the concept of consistency has been considered. However, their consistency modeling is mainly used to capture the individual-level correlation patterns (e.g., the consistency between detected objects' appearance and motion). In contrast, our work exploits the global spatial-temporal consistency of crowd motion and considers the multi-scale issue unique for crowd-level abnormal behaviors.

\noindent\textbf{Crowd Behavior Analysis.} Research on crowd behavior in public space has been an intriguing topic for the past decades~\cite{sanchez2020revisiting,luo2022why}. However, the analysis of crowd behaviors during crowd disasters is still very limited due to the scarcity of data. Based on the Hajj 2006 disaster videos, the local density, local velocity, and crowd pressure are measured to analyze the transitions from the normal crowd to stop-and-go wave and crowd turbulence~\cite{helbing2007dynamics}. Histograms of optical flow extracted from Love Parade 2010 disaster videos are used to cluster the motion patterns, and the magnitude and standard deviation of optical flow motion are combined to assess shock waves in crowd turbulence~\cite{krausz2012loveparade}. The temporal patterns in the Love Parade stampede are analyzed based on distance-based and point process representations of pedestrian movements using the extracted trajectories~\cite{lian2017long}. The measurements used in the above methods can capture certain aspects of abnormal behavior patterns. For instance, a high standard deviation of magnitude of optical flow is used to reflect the co-existence of moving people due to pushing and non-moving people in crowd turbulence~\cite{krausz2012loveparade}. However, other normal situations (e.g., visiting market) could also show high variation in speed. Thus, such measurements adopted to evaluate certain characteristics of anomalies are difficult to capture the full range of behavior patterns that are sufficient to distinguish CABs from all normal behaviors. In our work, we leverage the fine-grained feature learning capability of graph convolutional network (GCN) to comprehensively capture the patterns of correlation in crowd motion to effectively distinguish CABs from normal ones.

\section{Methodology}

In this section, we first present how the spatial and temporal features of crowd motion consistency are extracted from videos and represented in a graph form. Then, we introduce the multi-scale motion consistency network, denoted as \textbf{MSMC-Net}, which utilizes the multi-scale motion consistency information to train the network and detect CABs through attention-based unsupervised learning.


\subsection{Crowd Motion Consistency Representation}

To capture global features of crowd motion, we first present the measurements of the spatial consistency and temporal consistency, respectively. Then, we introduce a graph-based representation that encompasses both spatial and temporal consistency features at multiple scales.

\vskip 0.03in
\noindent\textbf{Spatial Crowd Motion Consistency.}
To measure the spatial consistency of crowd motion, we first extract the optical flow field~\cite{farneback2003two2} for every two consecutive frames and then divide the optical flow field of $H\times W$ area of the video frame into $h \times w$ number of regions. The optical flow vectors within each region are used to calculate an average velocity over each region. The average velocity field of this frame can be expressed as a matrix $\mathbf{M}_t = \{\mathbf{\bar{v}}_{t,c_i}\}_{i=1}^{h \times w}$, where $\mathbf{\bar{v}}_{t,c_i} $ represents the average velocity at region $c_i$ of frame $t$.

We propose to use spatial-inner consistency ($ \Omega^{\mathrm{sp}} $) to measure the uniformity degree within one region and spatial-inter consistency ($ \Gamma^{\mathrm{sp}} $) to measure the similarity of the average velocities of two adjacent regions. The spatial-inner consistency is used to capture the internal disorganization, such as the case of escape in different directions~\cite{zhao2019panic}. The spatial-inter consistency is used to evaluate the relationship between neighbor regions. For example, two opposite crowds in counter flows can produce the lowest spatial-inter consistency~\cite{crociani2017micro}.

The spatial-inner consistency is calculated using the spatial velocity entropy within a region.
Based on the optical flow vectors of a region, the vector direction space is discretized into $D$ classes: $\{\mathbf{v}_1,\mathbf{v}_2, \dots ,\mathbf{v}_D\}$, such as up, down, left, right, etc. $H_{t,c_i}^{\mathrm{sp}}(\mathbf{v}_p) $ is used to represent the number of optical flow vectors whose direction belongs to $\mathbf{v}_p$ at region $c_i$ of frame $t$. The distribution probability of optical vectors at region $c_i$ of frame $t$ can be computed by $P_{t,c_i}^{\mathrm{sp}}(\mathbf{v}_p)=H^{\mathrm{sp}}_{t,c_i}(\mathbf{v}_p) /n$, where $n$ is the total number of the pixel-level optical flow vectors at the region $c_i$ of frame $ t $. The spatial-inner consistency at region $ c_i $ of frame $ t $ can be calculated as follows:
\begin{equation}
\Omega^{\mathrm{sp}}_{t,c_i}=-\sum_{p=1}^{D} P^{\mathrm{sp}}_{t,c_i}(\mathbf{v}_p) \log P^{\mathrm{sp}}_{t,c_i}(\mathbf{v}_p).
\end{equation}

For the spatial-inter consistency, the difference in the average velocities of two regions is measured based on the adjusted cosine similarity, which measures the angular difference and the absolute value difference of vectors.
The spatial-inter consistency can be calculated as follows:
\begin{equation}
\Gamma^{\mathrm{sp}}_{t,(c_{i},c_{j})}  =  \cos (\mathbf{\bar{v}}_{t,c_i},\mathbf{\bar{v}}_{t,c_j})
(1-\frac{\left|\left\|\mathbf{\bar{v}}_{t,c_i}\right\|-\left\|\mathbf{\bar{v}}_{t,c_j}\right\|\right|}
{\left\|\mathbf{\bar{v}}_{t,c_i}\right\|+\left\|\mathbf{\bar{v}}_{t,c_j}\right\|}),
\end{equation}
where $ c_i $ and $ c_j $ are two adjacent regions and $\mathbf{\bar{v}}_{t,c_i},\mathbf{\bar{v}}_{t,c_j}$ are their average velocities at frame $t$.

\begin{figure*}[t]
\centering
\includegraphics[width=1.8\columnwidth]{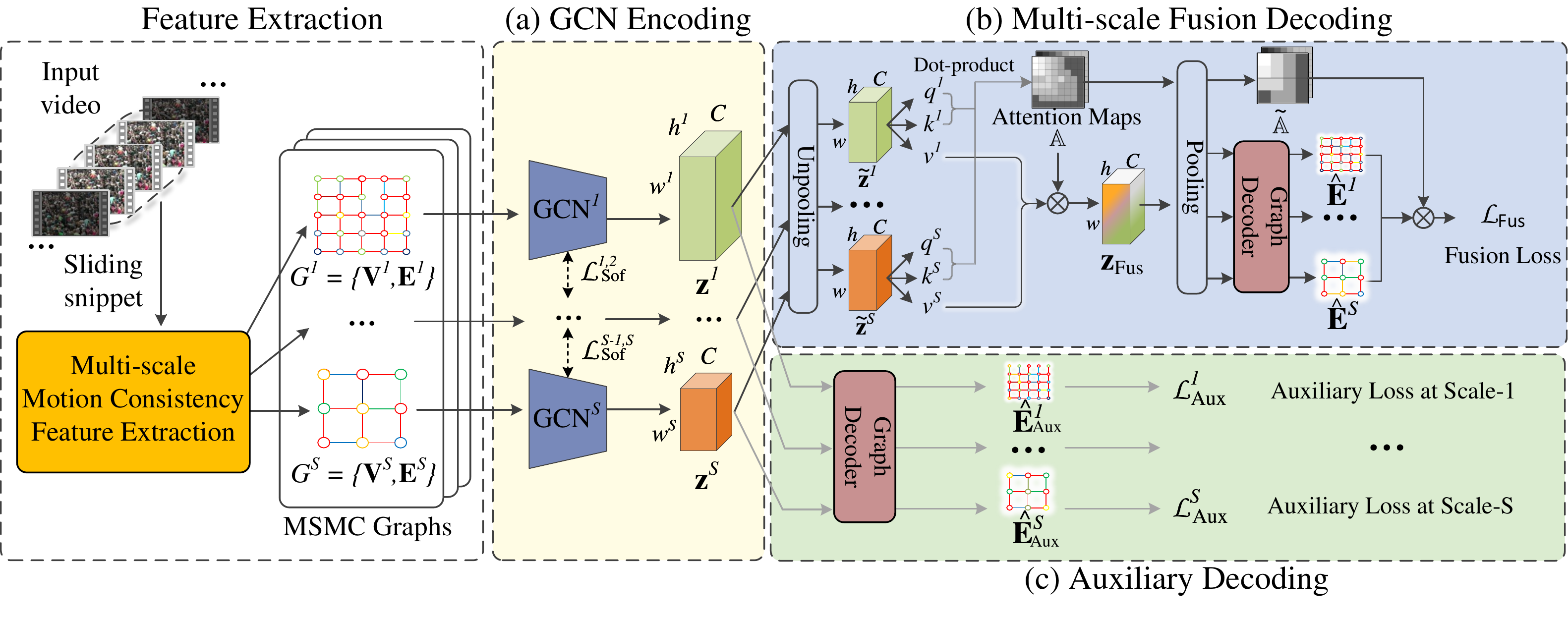}
\vskip -0.18in
\caption{
Overview of our MSMC-Net. Part (a) receives the extracted MSMC graphs to generate multi-scale embedding vectors. Two decoding processes in parts (b) and (c) are then performed to reconstruct the MSMC graphs for anomaly detection.
}\label{fig:architect}
\end{figure*}

\vskip 0.03in
\noindent\textbf{Temporal Crowd Motion Consistency.}
To measure the temporal consistency of crowd motion over a period of time, a sliding window approach is adopted to separate the given video of $T$ frames into sliding snippets. Each sliding snippet contains $m$ frames, and the window is slid by $\tau$ frames each time. An average velocity field sequence of the snippet starting with frame $t$ can be expressed as $\{ \mathbf{M}_{t}, ... ,\mathbf{M}_{t+m} \}$. Temporal features are then extracted over each snippet.

Similar to spatial consistency, we propose temporal-inner consistency ($ \Omega^{\mathrm{tp}} $) to measure the uniformity degree of one region's average velocity during its change over time and temporal-inter consistency ($ \Gamma^{\mathrm{tp}} $) to measure the uniformity degree of two adjacent regions' average velocities over time.
The temporal consistency compensates for the spatial consistency for obtaining time-varying features. For instance, the temporal-inner consistency can help detect the frequent velocity change over time when crowd turbulence starts~\cite{helbing2007dynamics,ma2013new}. The temporal-inter consistency can capture pedestrians' synchronized movement over time in turbulence regions~\cite{lian2016correlation}.

Based on the average velocity field sequence of a snippet starting with frame $t$, temporal velocity entropy is calculated to measure the temporal-inner consistency. Similar to the calculation of spatial-inner consistency, the average velocities are divided into $ D $ categories, and $ H^{\mathrm{tp}}_{t+m,c_i}(\mathbf{v}_p)$ measures the number of times, in which the direction of average velocity of region $c_i$ belongs to $\mathbf{v}_p$ during frame $t$ to $t+m$. The velocity distribution probability over this period can be computed by $P^{\mathrm{tp}}_{t+m,c_i}(\mathbf{v}_p) = H^{\mathrm{tp}}_{t+m,c_i}(\mathbf{v}_p) / m$, where $m$ is the length of the snippet. Temporal velocity entropy at region $ c_i $ of frame $ t $ is calculated as follows:
\begin{equation}
\Omega^{\mathrm{tp}}_{t+m,c_i}=-\sum_{p=1}^{D} P^{\mathrm{tp}}_{t+m,c_i}(\mathbf{v}_p) \log P^{\mathrm{tp}}_{t+m,c_i}(\mathbf{v}_p).
\end{equation}

To measure the temporal-inter consistency between two regions, we leverage the concept of mutual information to describe the correlations between two regions' motion over time. The distribution probability of two adjacent regions' velocities $  {P}^{\mathrm{tp}}_{t+m,c_i}(\mathbf{v}_p), {P}^{\mathrm{tp}}_{t+m,c_j}(\mathbf{v}_q) $ and
their joint distribution probability $ {P}^{\mathrm{tp}}_{t+m}(\mathbf{v}_p,\mathbf{v}_q) $ are first obtained. The temporal-inter consistency between region $c_i$ and $c_j$ from frame $t$ to frame $t+m$ can be calculated as follows:
\begin{small}
\begin{equation}
\Gamma^{\mathrm{tp}}_{t+m,(\!c_i,c_j\!)}\!=\!
\sum_{p,q}^{D\!,D}\!
 {P}^{\mathrm{tp}}_{t+m}(\!\mathbf{v}_p,\!\mathbf{v}_q\!) \!\log \!
\frac{{P}^{\mathrm{tp}}_{t+m}(\mathbf{v}_p,\!\mathbf{v}_q)}
{{P}^{\mathrm{tp}}_{t+m,c_i}\!(\!\mathbf{v}_p\!) {P}^{\mathrm{tp}}_{t+m,c_j}\!(\!\mathbf{v}_q\!)}\!.
\end{equation}
\end{small}

\vskip 0.03in
\noindent\textbf{Construction of Motion Consistency Graphs. }
To characterize the crowd motion consistency of frame $t$,
the spatial consistency feature of frame $t$ and the temporal consistency feature of the snippet ending with frame $t$ are used.
A graph structure capturing both spatial and temporal consistency information of snippet $t$ is proposed as follows:
$ G_{t}=\{ \mathbf{V}_{t}, \mathbf{E}_{t} \} $,
where each divided region in the frame is considered as a vertex, adjacent regions are connected by edges, and $ \mathbf{V}_{t} $ and $ \mathbf{E}_{t} $ represent the set of vertexes and edges at frame $t$, respectively.
Each vertex for a region $c_i$ contains the information of the spatial-inner consistency and temporal-inner consistency in the form of vectors:
$$ \mathbf{V}_{t,c_i} = [ \Omega^{\mathrm{sp}}_{t,c_i}; \Omega^{\mathrm{tp}}_{t,c_i}].  $$
The weight of the edge connecting two adjacent regions $c_i$ and $c_j$ contains the information of the spatial-inter consistency and temporal-inter consistency in the form of vectors:
$$ \mathbf{E}_{t,(c_i,c_j)} =
[ \Gamma^{\mathrm{sp}}_{t,(c_{i},c_{j})} ;
\Gamma^{\mathrm{tp}}_{t,(c_{i},c_{j})}]. $$
For a given video, a sequence of motion consistency graphs is generated, and each graph is generated per snippet. The first graph $G_m$ is constructed based on the first sliding snippet from frame 1 to $m$. The subsequent graphs are generated when the window of snippet is slid by $\tau$ frames.

However, using only a unified scale to analyze crowd motion makes it difficult to adapt to the variable range of crowd behaviors. Therefore, it is necessary to extract multi-scale crowd motion features for adaptive learning of crowd behaviors. To this end, we define the baseline scale using the average size of the pedestrians in a given video. The video frame of $W*H$ size is divided into $h^1 \times w^1$ number of regions, where $W/w^1 $ and $H/h^1 $ are ensured to be closest to the pedestrians' average shoulder pixel count as the baseline scale (1x-scale).
Based on $s$ times the baseline scale, we can perform a $s$x-scale division to obtain $\left \lceil \frac{h^1}{s}  \right \rceil   \times \left \lceil \frac{w^1}{s}  \right \rceil $ number of regions. The motion consistency graphs above can be extracted at different scales. The multi-scale motion consistency (MSMC) graphs at frame $t$ containing 1x-scale to $S$x-scale can be expressed as $ \{G_{t}^{s}\}_{s=1}^{S}$ and a sequence of MSMC graphs can then be generated for the entire video.

\subsection{Multi-scale Motion Consistency Network}
Based on the construction of the MSMC graphs described above, we aim to learn the behavioral patterns of the normal crowd under a proper scale level so that crowd-level abnormal behaviors can be detected by examining the deviation from the normal patterns. To this end, our MSMC-net is proposed to perform a multi-scale fusion-based reconstruction of the MSMC graphs in training and reconstruction-based anomaly detection in testing.

\vskip 0.03in
\noindent\textbf{Network Architecture.}
Figure~\ref{fig:architect} shows our proposed crowd motion learning framework, which consists of (a) GCN encoding, (b) multi-scale fusion decoding, and (c) auxiliary decoding. Part (a) receives the MSMC graphs extracted from the input video, and the GCN-based encoders~\cite{welling2016semi} are used to exploit the structural features in these graphs to capture the correlation in crowd motion. The GCN-based encoders produce a set of multi-scale embedding vectors. To solve the scale variation problem, part (b) takes the multi-scale embedding vectors and fuses them into a multi-scale fusion vector based on a self-attention mechanism. Fusion-based decoding is then performed to reconstruct the MSMC graphs. To prevent the fusion-based reconstruction from falling into a local optimum, part (c) introduce an auxiliary decoding process that reconstructs the MSMC graphs at each scale separately during training.

In part (a), the MSMC graphs $ \{G^{s}\}_{s=1}^{S} $ extracted from a given sliding snippet are encoded into multi-scale embedding vectors  $\{\mathbf{z}^s\}_{s=1}^S$. Since the edges in an MSMC graph are presented as two-dimensional vectors containing spatial and temporal consistency information, we adopt two GCNs for encoding each graph. One GCN aggregates spatial and temporal features of nodes using only spatial features of edges, and another uses only temporal features of edges. The embedding vectors from the two GCNs are concatenated to obtain the graph embedding vector $\mathbf{z}^s \in \mathbb{R}^{w^s \times h^s \times 2C}$ at each scale $s$, where $w^s \times h^s $ equals to the number of vertexes in the motion consistency graph of scale $s$ and $C$ is the embedding dimension of the encoder. The benefit of using two GCNs for multidimensional edge problem lies in its simple implementation and capability to tune GCN parameters separately for each dimension. For knowledge sharing between different scales, the reconstruction tasks are considered as a multi-task operation, and soft sharing constraints ($\mathcal{L}^{s_1,s_2}_{\mathrm{Sof}}$) are applied between the encoders at two different scales.

In part (b), the dimensions of multi-scale vectors are first reshaped to a unified one through nearest-neighbor unpooling. A self-attention mechanism is then leveraged to generate the self-attention maps representing each scale's relative importance. The multi-scale fusion process is performed via aggregating the reshaped vectors based on the attention maps, and a multi-scale fusion vector $\mathbf{z}_{\mathrm{Fus}}^{}$ is generated. Finally, the MSMC graphs are reconstructed with the multi-scale fusion vector, and a fusion loss ($\mathcal{L}_{\mathrm{Fus}}$) is defined to assess the multi-scale fusion-based reconstruction.

In part (c), the MSMC graphs are reconstructed at each scale using only the corresponding scale's embedding vectors obtained from part (a). For a given scale $s$, the decoder receives the embedding vector $\mathbf{z}^s$ at each scale $s$ and reconstructs the graph's edges ($\hat{\mathbf{E}}^s_{\mathrm{Aux}}$) via inner-product operation. This auxiliary reconstruction task helps to prevent the fusion-based reconstruction process in part (b) from falling into a local optimal solution. This is because if only fusion loss is used to perform both the scale selection and reconstruction learning, the scale that is preferred in scale selection will be more beneficial for reconstruction learning. This may lead to the optimization at a particular scale too much. The auxiliary loss ($\mathcal{L}_{\mathrm{Aux}}^{s}$) is thus used in part (c) to learn the reconstruction at each scale separately so as to prevent the neglect of some scales that do not perform well initially.

\vskip 0.03in
\noindent\textbf{Self-attention based Multi-scale Fusion.}
In different crowd scenarios, the scale of crowd behaviors can vary, which affects the quantification of crowd behavior patterns and detecting anomalies. To solve the scale variation problem, we leverage a self-attention mechanism, which is used to aggregate the motion consistency features at different scales to perform the multi-scale fusion-based reconstruction in part (b) of our MSMC-Net. During the reconstruction training process, the scale of the fusion model is automatically tuned such that the fusion loss (see in the next section) yields a minimum value.

To perform self-attention-based multi-scale fusion, the graph embedding vectors $\{\mathbf{z}^s\}_{s=1}^S$ are first unsampled across multiple scales to have the same size as the maximum scale through nearest-neighbor unpooling. The unified-scale vectors $ \{\tilde{\mathbf{z}}^{s}\}_{s=1}^{S}$ are obtained.
Then, the embedding vectors are used to guide the generation of attention maps through a cross-scale attention mechanism. The cross-scale attention map $ a_{xy}^{s} $ for scale $ s $ at position $ x,y $ of the map is calculated as follows:
\begin{equation}
\label{form:5}
a_{xy}^{s}=\frac{\left(W_{\mathrm{qry}}  \tilde{z}_{xy}^{s}\right)\left(W_{\mathrm{key}}  \tilde{z}_{xy}^{s}\right)^{T}}{\left\|W_{\mathrm{qry}}  \tilde{z}_{xy}^{s}\right\|\left\|W_{\mathrm{key}}  \tilde{z}_{xy}^{s}\right\|},
\end{equation}
where $ s \in\left\{1,2, \ldots, S\right\}$,
$x \in\{1,2, \ldots, w\}$,
$y \in\{1,2, \ldots, h\} $,
$ \|\cdot\| $ represents the norm function, $ W_{\mathrm{qry}}, W_{\mathrm{key}} $ are the trainable weight matrices for generating query ($q^s_{xy} = W_{\mathrm{qry}}  \tilde{z}_{xy}^{s}$) and key ($k^s_{xy} = W_{\mathrm{key}}  \tilde{z}_{xy}^{s} $) for scale $s$ at position $x,y$.
It can be further normalized to denote the current relative importance of scales:
\begin{equation}
\hat{a}_{xy}^{s}=\frac{\exp \left({a_{xy}^{s}}\right)}
{\sum_{s^{\prime}=1}^{S} \exp \left({a_{xy}^{s^{\prime}}}\right)}.
\end{equation}
The set of normalized attention maps for all the scales $\mathbb{A}=\{\mathbf{A}^{s}\}_{s=1}^{S}$ can be utilized as the weight for the fusion vector:
\begin{equation}
\mathbf{A}^{s}=
\begin{bmatrix}
   \hat{a}^s_{11}& \cdots  & \hat{a}^s_{w1} \\
   \vdots & \ddots & \vdots \\
  \hat{a}^s_{1h}& \cdots  & \hat{a}^s_{wh}
\end{bmatrix}
=\left [ \hat{a}_{xy}^{s}\right ],
\end{equation}
\begin{equation}
z_{\mathrm{Fus}}^{xy}=\sum_{s=1}^{S} \hat{a}_{xy}^{s} W_{\mathrm{val}} \widetilde{z}_{xy}^{s},
\end{equation}
where $ W_{\mathrm{val}} $ is the trainable weight matrix for generating value ($v^s_{xy} = W_{\mathrm{val}}  \tilde{z}_{xy}^{s}$) for scale $s$ at position $x,y$.

To achieve fusion-based reconstruction, the fusion vector $ \mathbf{z}_{\mathrm{Fus}}^{} = \left [ {z}^{xy}_{\mathrm{Fus}}\right ], $ and normalized attention map $\mathbb{A}$ are passed through pooling layers to reshape into the original scales. The reshaped fusion vectors $ \{\widetilde{\mathbf{z}}_{\mathrm{Fus}}^s\}_{s=1}^{S} $ are used for reconstructing the MSMC graphs. The set of reshaped attention maps $\widetilde{\mathbb{A}}=\{\widetilde{\mathbf{A}}^{s}\}_{s=1}^{S}$ is used to represent the importance of each scale for the aggregation of multi-scale fusion loss.

\vskip 0.03in
\noindent \textbf{Summary of reconstruction procedure.}
The reconstruction procedure of our MSMC-Net is shown in Algorithm~\ref{algo}.

\begin{algorithm}[ht]
\caption{Reconstruction procedure of our MSMC-Net}
\label{algo}
\textbf{Input}: MSMC graphs $ \{G^{s}\}_{s=1}^{S} $
\begin{algorithmic}[1] 
\FOR{$s=1 \to S$}
\STATE $\mathbf{z}^s = {\mathrm{GCN}}^s\left( G^s\right)$ \# GCN encoding
\STATE $\tilde{\mathbf{z}}^{s} = {\mathrm{Unpooling}}^s\left( \mathbf{z}^s\right)$  \# reshape into a unified scale
\STATE Obtain attention map $[a_{xy}^{s}]$ based on Eq.~\ref{form:5}
\ENDFOR

\STATE Obtain normalized attention maps $\mathbb{A}$ based on Eq.~6-7
\STATE Obtain fusion vector $\mathbf{z}_{\mathrm{Fus}}$ using Eq.~8

\FOR{$s=1 \to S$}
\STATE $\tilde{\mathbf{A}}^{s} = {\mathrm{Pooling}}^s\left( \mathbf{A}^s\right)$  \# reshape into original scales
\STATE $\tilde{\mathbf{z}}_{\mathrm{Fus}}^s = {\mathrm{Pooling}}^s\left( \mathbf{z}_{\mathrm{Fus}}\right)$
\STATE $\hat{\mathbf{E}}^{s}= {\mathrm{Decode}}\left(\tilde{\mathbf{z}}_{\mathrm{Fus}}^s \right)$ \# part (b) reconstruction
\STATE $\hat{\mathbf{E}}_{\mathrm{Aux}}^{s}= {\mathrm{Decode}}\left(\mathbf{z}^{s}\right)$ \# part (c) reconstruction
\ENDFOR

\STATE \textbf{return} $ \{\hat{\mathbf{E}}^{s}\}_{s=1}^{S} ,\{\hat{\mathbf{E}}_{\mathrm{Aux}}^s\}_{s=1}^{S} $ and $ \{\tilde{\mathbf{A}}^{s}\}_{s=1}^{S} $.
\end{algorithmic}
\end{algorithm}

\begin{table*}[th]
\centering
\scalebox{0.97}{
\begin{tabular}{l|c|c|c|c|c|c}
\hline
\multirow{2}{*}{\diagbox{method}{dataset}} & \multicolumn{2}{c|}{\textbf{UMN}} & \multicolumn{2}{c|}{\textbf{Hajj}} & \multicolumn{2}{c}{\textbf{Love Parade}}  \\
\cline{2-7}
                        & AUC ($\%, \uparrow$) & EER ($\%, \downarrow$)            & AUC ($\%, \uparrow$) & EER  ($\%, \downarrow$)             & AUC ($\%, \uparrow$) & EER ($\%, \downarrow$)                  \\
\hline
        AMC            & 85.4 $\pm$ 1.3 & 20.3 $\pm$ 1.3 & 65.3 $\pm$ 0.5  & 33.5 $\pm$ 0.8  & 59.6 $\pm$ 1.4  & 39.4 $\pm$ 1.2   \\
        FramePred      & 87.9 $\pm$ 1.7 & 19.6 $\pm$ 1.4 & 70.2 $\pm$ 1.3  & 32.2 $\pm$ 0.9  & 58.7 $\pm$ 2.2  & 40.9 $\pm$ 1.7  \\
        AMMC-Net       & 87.5 $\pm$ 2.3 & 20.3 $\pm$ 1.9 & 71.4 $\pm$ 1.8  & 30.1 $\pm$ 1.8  & 53.7 $\pm$ 1.6  & 45.4 $\pm$ 2.7  \\
        MNAD (recons)  & 85.1 $\pm$ 0.7 & 24.2 $\pm$ 1.1 & 82.4 $\pm$ 2.6  & 24.3 $\pm$ 1.8  & 57.6 $\pm$ 1.6  & 42.3 $\pm$ 2.1   \\
        MNAD (pred)    & 88.6 $\pm$ 0.4 & 19.3 $\pm$ 0.7 & 73.2 $\pm$ 0.7  & 31.8 $\pm$ 0.9  & 56.8 $\pm$ 1.1  & 46.9 $\pm$ 1.7   \\

\hline
        \textbf{MSMC-Net (Ours)}  &  \textbf{94.4 $\pm$ 0.5}  &  \textbf{12.1 $\pm$ 1.0}  &  \textbf{92.3 $\pm$ 0.5} &  \textbf{18.0 $\pm$ 1.8}   &  \textbf{82.2 $\pm$ 0.9} & \textbf{22.5 $\pm$ 0.5} \\
\hline
\end{tabular}
}
\caption{Frame-level detection results of our method and the compared baseline methods. All the results support the statistically significant improvement of our method over the baseline methods by a two-sample $t$-test at a 0.05 significance level.}
\label{table:cmp}
\vskip -0.1in
\end{table*}

\subsubsection{Loss Functions.}
In the training phase, the MSMC-Net is optimized by minimizing the following objective functions: fusion loss $\mathcal{L}_{\mathrm{Fus}}$, auxiliary losses $\mathcal{L}_{\mathrm{Aux}}$ and soft sharing losses $\mathcal{L}_{\mathrm{Sof}}$.
\begin{equation}
\mathcal{L} = \lambda_{\mathrm{Fus}}\mathcal{L}_{\mathrm{Fus}}  + \lambda_{\mathrm{Aux}}\sum_{s=1}^{S} \mathcal{L}^{s}_{\mathrm{Aux}} + \lambda_{\mathrm{Sof}}\sum_{s_{1},s_2}^{S,S}
\mathcal{L}_{\mathrm{Sof}}^{s_1,s_2},
\end{equation}
where the hyper-parameters $ \lambda_{\mathrm{Fus}} $, $ \lambda_{\mathrm{Sof}} $, $ \lambda_{\mathrm{Aux}} $ are used to tune the importance of each part.

To achieve multi-scale fusion-based reconstruction, the reshaped attention map $\widetilde{\mathbb{A}}$ is utilized to weigh different scales' losses to obtain the fusion loss as follows:
\begin{equation}
\mathcal{L}_{\mathrm{Fus}}
\!=\!
\sum_{s=1}^{S}
\frac{1}{w^{s}\! \times \!h^{s}}\!
\sum_{x_i y_i,x_j y_j}\!
\widetilde{a}_{x_i y_i}^{s}
\widetilde{a}_{x_j y_j}^{s}
\mathcal{L}_{(x_i y_i,x_j y_j)}^{s},
\end{equation}
where $\mathcal{L}_{(x_i y_i,x_j y_j)}^{s}=\!\left\|\mathbf{E}_{(x_i y_i,x_j y_j)}^{s}-\hat{\mathbf{E}}_{(x_1 y_1,x_2 y_2)}^{s}\right\|$ denotes the $\ell_2$ distance between the fusion-based reconstructed edge and the original edge at scale $s$.

To avoid the pitfall in multi-scale fusion-based decoding as described previously, the auxiliary loss is used in auxiliary decoding, which is defined as the $\ell_2$ distance between the auxiliary reconstructed edges and the original at each scale $s$ as follows:
\begin{equation}
\mathcal{L}_{\mathrm{Aux}}^{s}=
\left\|\mathbf{E}^{s}-\hat{\mathbf{E}}^s_{\mathrm{Aux}}\right\|,
\end{equation}
where $\hat{\mathbf{E}}^{s}_{\mathrm{Aux}}$ denotes the reconstructed edges via auxiliary decoding at scale $s$.

To share knowledge between the GCN encoders at two different scales, the soft sharing loss is defined as:
\begin{equation}
\mathcal{L}^{s_1,s_2}_{\mathrm{Sof}}=\left\|W^{s_{1}}-W^{s_{2}}\right\|,
\end{equation}
where $W^{s_{1}},W^{s_{2}}$ are the sets of parameters of the encoders at scale $s_1$ and $s_2$ , respectively.

\vskip 0.03in
\noindent \textbf{Fusion-based Anomaly Score.}
In the testing phase, given the learned normal correlation patterns of crowd motion, anomalies are detected by examining the deviation from these normal correlation patterns. Since the scale of crowd behavior has been learned during the training phase, only the multi-scale fusion loss $ \mathcal{L}_{\mathrm{Fus}} $ is used for detecting anomaly in the testing phase. Considering that crowd-level abnormal behaviors tend to last for a period of time, the moving average is adopted to obtain the anomaly score $\mathcal{S}_{t}$ at each frame $t$ as follows:
\begin{equation}
\mathcal{S}_{t} = (1-\lambda_{\mathrm{Mov}}) \times \mathcal{S}_{t-1} +\lambda_{\mathrm{Mov}} \times \mathcal{N}(\mathcal{L}_{\mathrm{Fus}}),
\end{equation}
where $ \lambda_{\mathrm{Mov}} $ is the weight of moving average, $ \mathcal{N}  (\cdot ) $ denotes the min-max normalization, $ \mathcal{S}_{t-1} $ is the anomaly score in the previous frame and $\mathcal{S}_{0}\! = \!\mathcal{N}(\mathcal{L}_{\mathrm{Fus}})$. The anomaly score is in the range of 0 to 1, and the score is closer to 1 indicating a higher degree of anomaly.

\section{Experiments}
\subsection{Datasets}

\noindent We evaluate the performance of our method on three publicly available datasets, \textbf{UMN}, \textbf{Hajj} and \textbf{Love Parade}, which contain crowd-level abnormal behaviors including crowd escaping, counter flow and crowd turbulence. \textbf{UMN} consists of walking and escaping captured by CCTV cameras in three wild scenes. \textbf{Hajj} is derived from surveillance videos of the annual religious pilgrimage in Saudi Arabia. \textbf{Love Parade} contains surveillance videos of the 2010 Love Parade crowd disaster. Note that our work is the first to introduce \textbf{Hajj} and \textbf{LoveParade} for VAD study. More details on these datasets are described in Appendix.


\subsection{Baseline Methods}

\noindent Four state-of-the-art unsupervised VAD methods are considered as baselines, which can be categorized as \emph{prediction} and \emph{reconstruction}-based methods. Two prediction-based methods include \textbf{AMC}~\cite{nguyen2019anomaly} and \textbf{FramePred}~\cite{liu2018future}. Two reconstruction-based methods include  \textbf{AMMC-Net}~\cite{cai2021appearance} and \textbf{MNAD}~\cite{park2020learning}. Note that \textbf{MNAD} also offers a variant of prediction-based version, and we evaluate both versions of \textbf{MNAD} (recons) and \textbf{MNAD} (pred). Details on these baseline methods and our settings are in Appendix.


\begin{table*}[ht]
\noindent
\centering
\hskip -0.11in
\scalebox{0.96}{
\begin{tabular}{l|l|c|c|c|c|c|c}
\hline
\multirow{2}{*}{} & \multirow{2}{*}{Ablated Method Spec} & \multicolumn{2}{c|}{\textbf{UMN}}   & \multicolumn{2}{c|}{\textbf{Hajj}} & \multicolumn{2}{c}{\textbf{Love Parade}}  \\
\cline{3-8}
        &      & AUC ($\%,\uparrow$) & EER ($\%,\downarrow$)   & AUC ($\%,\uparrow$) & EER ($\%,\downarrow$)  & AUC ($\%,\uparrow$) & EER ($\%,\downarrow$)   \\
\hline
\multirow{2}{*}{\parbox{1.0cm}{Consistency \\modules}} & \textbf{Spatial} + Multi-scale     & 88.4 $\pm$ 3.7 &     17.5 $\pm$ 2.8 &    73.6 $\pm$ 1.1 & 38.1 $\pm$ 0.7 & 51.6 $\pm$ 2.1 & 47.9 $\pm$ 0.5  \\
& \textbf{Temporal} + Multi-scale    & 91.6 $\pm$ 1.9  &     19.1 $\pm$ 3.8 &    86.0 $\pm$ 0.8 & 24.8 $\pm$ 1.0 & 80.3 $\pm$ 1.7 & 25.1 $\pm$ 2.6 \\
\hline
\multirow{3}{*}{Single scale} & Spatial+Temporal (\textbf{1x})    & 89.6 $\pm$ 0.8         & 20.6 $\pm$ 0.9        & 90.5 $\pm$ 1.1        & 22.5 $\pm$ 0.4          & 67.6 $\pm$ 0.2        & 34.6 $\pm$ 0.3  \\
& Spatial+Temporal (\textbf{2x})     & 73.2 $\pm$ 1.4         & 32.1 $\pm$ 2.1         & 84.6 $\pm$ 0.4        & 27.5 $\pm$ 1.3        & 74.9 $\pm$ 0.3         & 32.3 $\pm$ 0.7 \\
& Spatial+Temporal (\textbf{4x})     & 68.6 $\pm$ 1.8         & 33.1 $\pm$ 2.0       & 72.3 $\pm$ 0.6         & 38.1 $\pm$ 0.7       & 76.9 $\pm$ 0.5         & 30.2 $\pm$ 0.2  \\
\hline
Our full method & $\surd$   & \textbf{94.4 $\pm$ 0.5}  & \textbf{12.1 $\pm$ 1.0} & \textbf{92.3 $\pm$ 0.5} & \textbf{18.0 $\pm$ 1.8} & \textbf{82.2 $\pm$ 0.9} & \textbf{22.5 $\pm$ 0.5} \\
\hline
\end{tabular}
}
\caption{Ablation study on (1) crowd motion representation and (2) multi-scale fusion learning in our proposed method.}
\label{ablation}
\end{table*}

\subsection{Evaluation Metrics}
\noindent We use two commonly adopted metrics for the evaluation of frame-level anomaly detection.
\textbf{AUC} measures the \emph{area under ROC curve}, and the value is larger the better ($\uparrow$). \textbf{EER} measures the \emph{equal error rate} in terms of the location on ROC curve, where the false acceptance rate and false rejection rate are equal, and the value is smaller the better ($\downarrow$).

\subsection{Benchmark Results}
\noindent In our evaluation, ten independent training and testing runs for each method are performed. The average results of different VAD methods in terms of AUC and EER are shown in Table~\ref{table:cmp}.
It can be seen that our model achieves the highest average AUC indicating the best overall performance, and the lowest EER indicating that our method generates fewer false and missed alarms. It can also be seen that different datasets exhibit different complexities for anomaly detection. In the Love Parade dataset that contains crowd turbulence, our proposed method exceeds the existing methods significantly. This is probably because pedestrians are packed tightly and move coherently in crowd turbulence, which makes it more difficult to distinguish from the normal congested crowd by their appearance or motion. However, our method can discover its difference by examining both spatial and temporal crowd motion consistency. It should also be noted that the original UMN videos contain a text tag whenever an anomaly appears in the video (see Appendix), which unnecessarily makes the detection task easier. To prevent the influence of the tag, the UMN results in Table~\ref{table:cmp} are based on the trimmed videos removing those tags.

\begin{figure}[t] 
\centering
\vskip -0.1in
\includegraphics[width=0.48\textwidth]{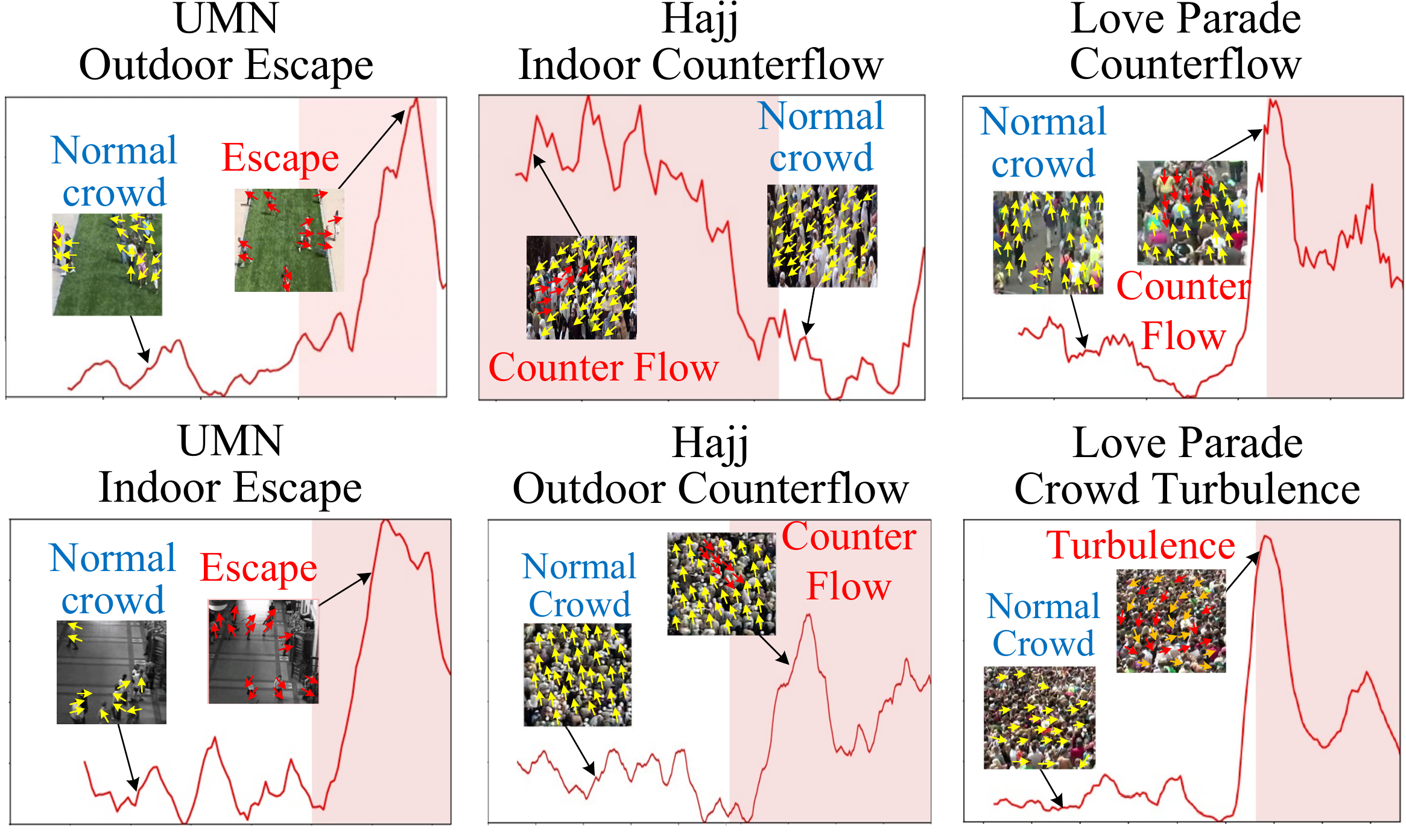}
\caption{
Frame-level anomaly score of our MSMC-Net on three datasets in the testing phase. Each pink area indicates the time interval that the anomaly occurs, and the red curve shows how the anomaly score changes over time.}
\label{cmp}
\vskip -0.1in
\end{figure}

In Figure~\ref{cmp}, we show how our anomaly scores change over time. It can be observed that our method can effectively produce low anomaly scores for normal situations and high ones for abnormal situations containing various CABs. The changes in anomaly score also reflect how crowd-level abnormal behavior evolves over time. For instance, in UMN, the anomaly score is low at the very beginning of the crowd escape event. It is because only pedestrians' directions change when they just start escaping, while their speed changes take a while due to inertia. The corresponding anomaly score is thus low when escape starts and gradually increases, as shown in Figure~\ref{cmp}. In Love Parade, the anomaly score fluctuates in crowd turbulence, since pedestrians are sometimes pushed to move or halted in crowd waves. More results on the evolution of anomaly scores in training and average running time in testing are in Appendix.



\subsection{Ablation Study}

We present the ablation study on crowd motion representation and multi-scale learning as shown in Table~\ref{ablation}.
\vskip 0.02in
\noindent \textbf{Effectiveness of Motion Consistency Representation.}
Two variants of our model are trained using the spatial and temporal motion consistency features to construct MSMC graphs, respectively. It can be seen that our full method gives the best results indicating that both the spatial and temporal motion consistency are useful for detecting CABs. It can also be seen that the model with only the temporal consistency feature outperforms the one with only the spatial consistency feature. This indicates that the unusual time-varying patterns occur more frequently than the unusual instantaneous patterns in some CABs. For instance, pedestrians in crowd turbulence frequently change their speeds, causing incongruence in crowd motion over time, whereas instantaneous patterns show subtle differences from the high-density normal crowd. However, the model with only the temporal consistency feature is not enough to achieve the best result. Adding spatial consistency can bring an AUC improvement of 2.8\%, 6.3\%, and 1.9\% for the three datasets.

\vskip 0.02in
\noindent\textbf{Effectiveness of Multi-scale Learning.}
To validate the significance of multi-scale learning, three single-scale variants were trained using motion consistency graphs constructed at 1x, 2x, and 4x scales, respectively. The reconstruction loss at a single scale as defined in our auxiliary loss is used for each ablated model. It can be observed that the scales achieving the best detection result are different for the three datasets. This demonstrates that the scale of crowd behaviors indeed varies for different scenarios. The 1x-scale single-scale model achieves the best result in UMN and Hajj, whereas the 4x-scale model performs best in Love Parade. A larger scale is required for Love Parade probably because the crowd turbulence patterns emerge from the interactions of many pedestrians in the high-density crowd. Moreover, the best scale for detecting CABs can change slightly over time for a given scenario. For example, it is found that the vortex size of crowd turbulence changes as the crowd moves~\cite{ivancevic2012turbulence}. Since our multi-scale fusion mechanism can adaptively estimate the proper scale over time when continuously receiving the video input, our multi-scale method can thus achieve better results than any single-scale variant, demonstrating the ability of scale adaptation.

\section{Conclusion}

This paper proposes a new VAD method for detecting crowd-level abnormal behaviors, which have not yet been fully investigated in existing VAD studies. The proposed method exploits the macroscopic crowd motion patterns via multi-scale motion consistency learning. A multi-scale motion consistency network is introduced to learn crowd motion features extracted at multi-scale and enable our model to estimate the proper scale of crowd behavior for detecting CABs. Evaluations based on real-world datasets demonstrate that our approach outperforms the state-of-the-art VAD methods for detecting CABs. Our future work may include validation on more real-world crowd event datasets and evaluation of cross-scenario detection performance.



\section*{Ethical Statement}

This work involves the use of crowd videos containing human subjects. However, the proposed method does not rely on the extraction of individual identity (e.g., face recognition) but rather the motion of the crowd for anomaly detection. Since the optical flow based on the tracking of moving pixels (e.g., head area) is used as our primary input, the proposed method should work even if the human faces in the video are obfuscated. Thus, privacy can be preserved to some extent.

\section*{Acknowledgements}
This work was supported in part by the National Natural Science Foundation of China under Grant 61872282, in part by the Key Research and Development Program of Shaanxi under Program 2022KW-06, in part by the National Natural Science Foundation of China under Grant U22A2035, and in collaboration of the project supported by MoE Tier 1 Grant RG12/21 (Singapore).

\bibliography{aaai23}

\clearpage
\appendix

\section{Appendix}\label{appendix}

\noindent This supplementary material is organized as follows. First, we describe the details of three crowd video datasets, including how the datasets are labeled and used for evaluation. Then, we introduce the details of the baseline methods used in the comparison, followed by the implementation details for our method and all the baseline methods. Lastly, we show the evolving anomaly detection score curves through training epochs and the running time evaluation.

\subsection{Datasets}
\label{appendix:dataset}

\begin{figure}[t]
\centering
    \begin{subfigure}[t]{0.47\textwidth}
           \centering
           \includegraphics[width=\textwidth]{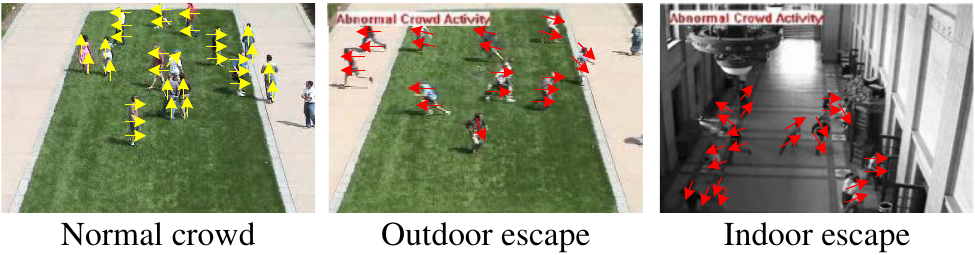}
            \caption{Example screenshots in UMN dataset }
            \label{fig:4a}
    \end{subfigure}
    \begin{subfigure}[t]{0.47\textwidth}
           \centering
           \includegraphics[width=\textwidth]{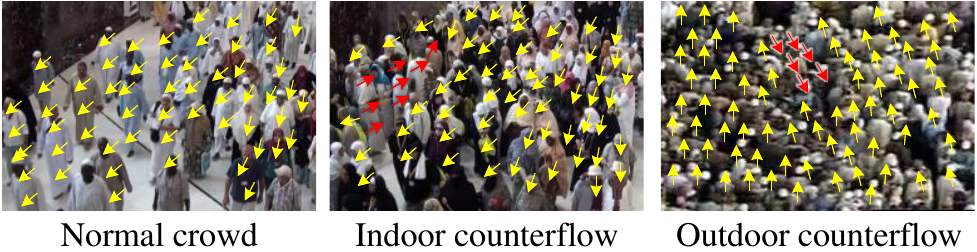}
            \caption{Example screenshots in Hajj dataset}
            \label{fig:4b}
    \end{subfigure}
    \begin{subfigure}[t]{0.47\textwidth}
           \centering
           \includegraphics[width=\textwidth]{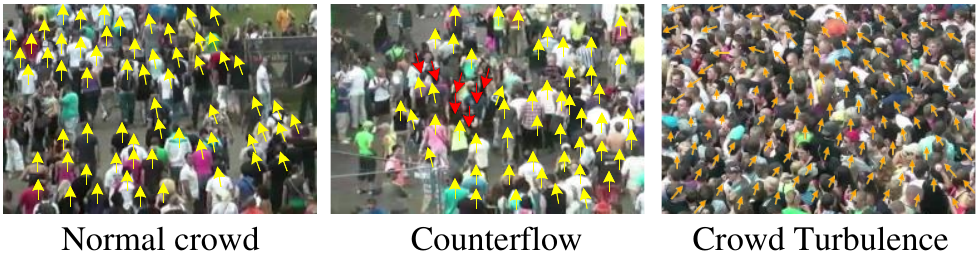}
            \caption{Example screenshots in Love Parade dataset}
            \label{fig:4c}
    \end{subfigure}

    \caption{
Example screenshots in our benchmark datasets. The arrows are added to indicate the optical flow in the frame. It is worth noting that during anomalies in the UMN dataset, there is a flashing text label in the upper left corner (see screenshots of outdoor and indoor escape), which is trimmed in our experiments to avoid its influence.
}
\label{fig:data}
\end{figure}

\noindent \underline{\textbf{UMN}}\footnote{\url{http://mha.cs.umn.edu/Movies/Crowd-Activity-All.avi}}
is a video shot by CCTV at the University of Minnesota in 2006 (see Figure~\ref{fig:4a}). The video contains three scenes. The video areas of the three scenes are 6 x 10 meters, 7 x 12 meters, and 10 x 9 meters, respectively. In the video, the walking pedestrians are considered normal, while the crowd escaping is abnormal. The original resolution of UMN is 320x240. As shown in Figure~\ref{fig:4a}, a flashing text label appears whenever a frame is regarded as abnormal. To avoid the influence of these labels on the anomaly detection task, we trim off the text label area of all video frames, and the trimmed video has a resolution of 320x213. The dataset is divided into non-overlapping training and testing parts. Our training set contains 4410 frames of normal behaviors, and our testing set contains 3300 frames of both normal (1823 frames) and abnormal behaviors (1477 frames). For labeling the ground truth, we use the original label already provided in the dataset.

\vskip 0.1in
\noindent \underline{\textbf{Hajj}}\footnote{\url{https://github.com/KAU-Smart-Crowd/Hajj_abnormal_behavior_detection}} comes from the surveillance video of Saudi Arabia's annual religious pilgrimage, including nine 45-second segments of dense crowds (see Figure~\ref{fig:4b}). In these videos, abnormal behaviors include standing, sitting, sleeping, running, moving in the opposite or different direction of the crowd, and non-pedestrian movements, such as cars and wheelchairs. Among these abnormal behaviors, we select the crowd-level abnormal behavior, namely counter flow, for evaluation. The resolution of the videos is 720x576. The selected dataset is divided into non-overlapping training and testing parts. The training set contains 1500 frames without counter flow behavior and the testing set contains 1080 frames with both counter flow behavior (480 frames) and normal behavior (600 frames). For labeling the ground truth, we use the original label already provided in the dataset.

\vskip 0.1in
\noindent\underline{\textbf{Love Parade}}\footnote{\url{https://loveparade2010doku.wordpress.com/2010/08/30/lopavent-veroffentlicht-originalvideos-von-7-der-16-uberwachungskameras-der-loveparade-2010/}} contains surveillance videos from 7 monitoring locations within 3 hours before the love parade accident in 2012 (see Figure~\ref{fig:4c}). The total length of all videos is over 23 hours and the crowd density in the videos reaches 11 people per square meter. The videos contain various CABs, including counter flow and crowd turbulence. We chose the videos from two of the cameras (4 and 13) in which counter flow and turbulence occurred and selected the video frames that contain the anomalies and are before and after the occurrence of anomalies. The selected dataset is divided into non-overlapping training and testing parts. The training set contains 2767 frames of normal behaviors and the testing set contains 1830 frames in total, in which 810 frames contain anomalies. For labeling the ground truth of the counter flow, frames are labeled based on whether the groups of pedestrians move in a counter direction from the majority of the crowd. To label the ground truth of the crowd turbulence, we refer to the post-disaster analysis of the Love Parade disaster and label the crowd turbulence events according to the provided timeline~\cite{helbing2012crowd}.
The first testing video containing the crowd turbulence is clipped from 14:50 to 15:10 of the Camera\_13\_1620 video, and the turbulence starts at 15:00. The second testing video containing the crowd turbulence is clipped from 13:50 to 14:11 of Camera\_04\_1440 video, and the turbulence starts at 14:04.

\subsection{Baseline Details}\label{appendix:baselines}

\noindent \underline{\textbf{FramePred}} \cite{liu2018future} is a prediction-based method that predicts the next frame with constraints in terms of appearance
(intensity loss and gradient loss) and motion (optical flow loss). It adopts U-Net as a generator to predict the next frame. It adopts the constraints of appearance (intensity loss and gradient loss) to generate a high-quality image and motion (optical flow loss) and detects anomaly by predicting future frames. Experimental results show that the AUC of the algorithm on CUHK Avenue, UCSD ped1, UCSD ped2, and ShanghaiTech datasets are 0.851, 0.831, 0.954, and 0.728, respectively.

\vskip 0.03in
\noindent \underline{\textbf{AMC}} \cite{nguyen2019anomaly} is a reconstruction-based method that infers anomaly scores from reconstruction errors of autoencoders. Anomaly detection is realized by combining Conv-AE and CNN of U-Net to reconstruct the frame, then comparing the difference between the reconstructed frame and the actual frame based on the patch scheme. Experimental results show that the AUC of the algorithm on CUHK Avenue and UCSD ped2 datasets are 0.869 and 0.962, respectively.

\vskip 0.03in
\noindent \underline{\textbf{MNAD (recons/pred)}} \cite{park2020learning} is a memory-based
method with both prediction and reconstruction-based variants. It exploits multiple prototypes to consider the various behaviors of normal data and uses a memory module to record the prototypical patterns of the items in the memory. At the same time, two modes of anomaly detection, prediction, and reconstruction, are proposed. Based on the memory module, novel feature compactness and separability loss are proposed to train memory, which ensures the diversity and discrimination of memory items. Experiments show that the AUC values of the reconstruction-based scheme on UCSD ped2, CUHK Avenue, and ShanghaiTech datasets are 0.902, 0.828, and 0.698, respectively; The AUC of the prediction-based scheme is 0.970, 0.885 and 0.705, respectively.

\vskip 0.03in
\noindent \underline{\textbf{AMMC-Net}} \cite{cai2021appearance} is a prediction-
based method, which leverages the prior knowledge of appearance and motion signals to capture the relationship between them in high-level features. The method fully uses the prior knowledge of appearance and motion signals and captures the corresponding relationship between them in the high-level feature space. Then, combined with multi-view features, a more fundamental and robust feature representation of conventional events is obtained, which can significantly increase the gap between abnormal and normal events. Experimental results show that the AUC of the algorithm on UCSD ped2, CUHK Avenue, and Shanghai tech datasets are 0.966, 0.866, and 0.737, respectively.

\begin{table}[t!]
    \centering
    \resizebox{\linewidth}{!}{
    \begin{tabular}{ll}
        \hline
        Parameters                                         & Value (UMN, Hajj, LP) \\
        \hline

        self-attention dimension
        ($q,k,v$)                      & 32, 16, 32 \\
        graph embedding dimension($z$)  & 32, 16, 32 \\
        encoder network size  &     (32, 16), (16,8), (32, 16)\\
        learning rate                   & $ 3e^{-4} $, $ 1e^{-4} $, $ 3e^{-4} $ \\
        \hline
        $ \beta_{1} $ of Adam optimizer              & 0.9     \\
        $ \beta_{2} $ of Adam optimizer              & 0.999   \\
        \hline
        velocity direction categories ($D$)            & 8     \\
        length of sliding window ($m$)              & 20       \\
        sliding window step ($\tau$)                        & 1  \\
        multi-scale number ($S$)                  & 3  \\
        fusion loss weight ($ \lambda_\mathrm{Fus} $)      & 1 \\
        soft sharing loss weight ($ \lambda_\mathrm{Sof} $)    & 1 \\
        auxiliary loss weight  ($ \lambda_\mathrm{Aux} $)    & 1 \\
        moving average weight ($ \lambda_\mathrm{Mov} $)  & 0.2 \\
        random seed                                  & 42 \\
        \hline
    \end{tabular}
    }
    \caption{Parameter settings}
    \label{hyper}
    \vskip -0.1in
\end{table}

\begin{figure*}[t]
\centering
\begin{subfigure}[t]{0.99\textwidth}
           \centering
           \includegraphics[width=0.99\textwidth]{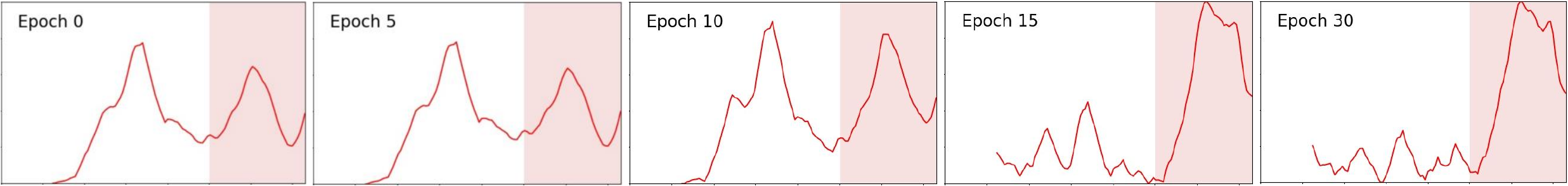}
            \caption{UMN}
            \label{evo:a}
    \end{subfigure}
    \begin{subfigure}[t]{0.99\textwidth}
            \centering
            \includegraphics[width=0.99\textwidth]{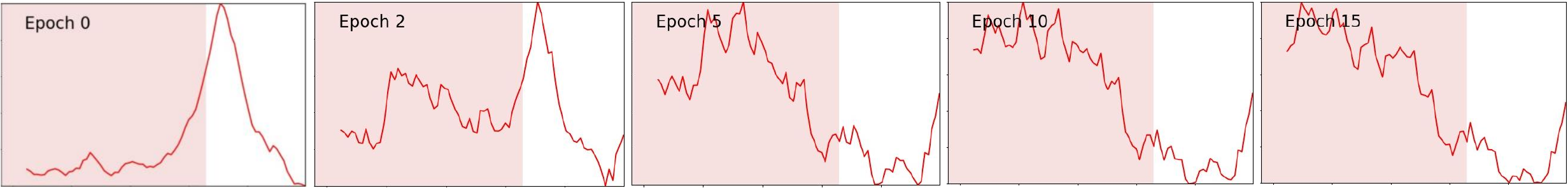}
            \caption{Hajj}
            \label{evo:b}
    \end{subfigure}
		\begin{subfigure}[t]{0.99\textwidth}
            \centering
            \includegraphics[width=0.99\textwidth]{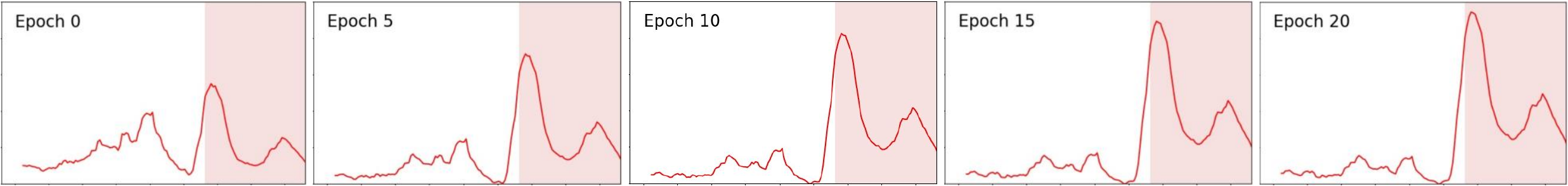}
            \caption{Love Parade}
            \label{evo:c}
    \end{subfigure}
\caption{
Evolution of anomaly score in the testing data as our MSMC-Net is trained over epochs. Each pink area indicates the time interval that the anomaly occurs and the red curve shows how the anomaly score changes over time.
}
\label{evo}
\end{figure*}

\vskip -0.03in
\subsection{Implementation Details}\label{appendix:implementation}
All our experiments are conducted on NVIDIA Tesla T4 GPUs with Intel(R) Xeon(R) 6230 2.10 GHz CPUs and 32 GB RAM operated with a 64-bit Windows 10 system.

\noindent \underline{\textbf{MSMC-Net (Ours)}}
We use a grid search to set hyper-parameters on the test split, and our hyper-parameters tuned for each dataset can be found in Table~\ref{hyper}.
The decay rates ($ \beta_{1} $,$ \beta_{2} $) of the Adam optimizer~\cite{kingma2015adam} for network parameter tuning are set to 0.9 and 0.999, as suggested in the original paper. The number of velocity direction categories ($D$) is set to 8, representing top, bottom, left, right, top left, bottom left, etc. Considering a moderate range of scales, our multi-scale framework is tested using the scale range of \{1x, 2x, 4x\} scales. To train all parts in a balanced way, we set an equal weight for the fusion loss, auxiliary loss, and soft sharing loss. Considering crowd behaviors tend to last for a period, the weight of the moving average ($ \lambda_\mathrm{Mov} $) is set to 0.2. The region size $w\times h$ of the baseline $1x$ scale for each dataset is determined by the average pixel size of pedestrians in the dataset's videos, as described in the Methodology section. The random seed is set to 42 when initializing the network's parameters using the random seed function provided in Python. Farneback optical flow~\cite{farneback2003two2} is obtained using OpenCV\footnote{\url{https://docs.opencv.org/2.4/modules/video/doc/motion_analysis_and_object_tracking.html?highlight=calcopticalflowfarneback}}. All codes are implemented in Python, and our MSMC-Net is trained using PyTorch\footnote{\url{https://pytorch.org/}}.

\vskip 0.03in
\noindent \underline{\textbf{FramePred}}~\cite{liu2018future} The code is obtained from the link\footnote{\url{https://github.com/StevenLiuWen/ano_pred_cvpr2018}} and a grid search is used to set the hyper-parameters on the same test split as used for tuning our method. The learning rates are set to ($2e^{-4}$, $2e^{-5}$), ($1e^{-4}$, $1e^{-5}$) and ($2e^{-4}$, $2e^{-5}$) on the three datasets respectively. As stated in the original paper, $\lambda_{int}$, $\lambda_{gd}$, $\lambda_{op}$ and $\lambda_{adv}$ vary from different datasets very slightly and they are set to 1.0, 1.0, 2.0 and 0.05, respectively. For a fair comparison, the sliding window size is set to 20, the same as in our method.

\vskip 0.03in
\noindent \underline{\textbf{AMC}}~\cite{nguyen2019anomaly}
The code is obtained from the link\footnote{\url{https://github.com/nguyetn89/Anomaly_detection_ICCV2019}} and a grid search is used to set the hyper-parameters on the same test split as used for tuning our method. The learning rates are set to ($1e^{-4}$, $1e^{-5}$), ($5e^{-5}$, $5e^{-6}$) and ($1e^{-4}$, $1e^{-5}$) on the three datasets, respectively.

\vskip 0.03in
\noindent \underline{\textbf{MNAD}}~\cite{park2020learning}
The code is obtained from the link\footnote{https://github.com/cvlab-yonsei/MNAD} and a grid search is used to set the hyper-parameters on the same test split as used for tuning our method. For the reconstruction version of MNAD, the learning rates are set as $2e^{-5}$, $1e^{-5}$ and $2e^{-5}$, and $\lambda$ is set to 0.9, 0.7 and 0.6 for the three datasets, respectively. For the prediction version of MNAD, the learning rates are set as $2e^{-4}$, $1e^{-4}$ and $2e^{-4}$, and $\lambda$ is set to 0.9, 0.5 and 0.5 for the three datasets, respectively. As for the height $H$ and width $W$ of the query feature map, the number of feature channels $C$ and memory items $M$ hardly vary among different datasets. They are set to 32, 32, 512 and 10, as in the original paper. The sliding window size is set to 20, the same as in our method for a fair comparison.

\vskip 0.03in
\noindent \underline{\textbf{AMMC-Net}}~\cite{cai2021appearance}
The code is obtained from the link\footnote{\url{https://github.com/NjuHaoZhang/AMMCNet_AAAI2021}} and a grid search is used to set the hyper-parameters on the same test split as used for tuning our method. The learning rates are set as $1e^{-3}$, $3e^{-4}$ and $1e^{-3}$ for the three datasets. According to the ablation study in the original paper, the memory item numbers $K$, memory size $N$ and memory dimension $D$ are set to 2, 64 and 256. For a fair comparison, the sliding window size is also set to 20, the same as in our method.

\vskip -0.03in
\subsection{Evolving Anomaly Scores}\label{appendix:curves}
\vskip -0.03in
Figure~\ref{evo} shows how the anomaly score in our testing data is improved as the training of our MSMC-net is performed over epochs. As the training proceeds, it can be seen that our method can generate a higher score for abnormal behavior and a lower score for normal behavior.

%
%
%
%
%
%
%

\vskip -0.03in
\subsection{Running Time}\label{appendix:running}
Given the hardware configuration as stated previously, the average running time of our trained model is approximately 75 ms, which contains crowd motion feature extraction (45 ms) and neural network computation (30 ms). The average running times of baseline methods, including AMC, FramePred, AMMC-Net, MNAD(recons), and MNAD(pred), are 81ms, 160ms, 55 ms, 34ms, and 61 ms, respectively, in our experiments. In overall, the average running time for executing our trained model to detect CABs in the testing videos is comparable to the baseline methods.

\end{document}